\documentclass{article}
\usepackage{log_2022}   

\usepackage{booktabs}           
\usepackage{multirow}           
\usepackage{amsfonts}           
\usepackage{graphicx}           
\usepackage{duckuments}         
\usepackage{enumitem}
\usepackage{xcolor}
\usepackage{algorithmic}
\usepackage{subfigure}
\usepackage{booktabs}
\usepackage{amsmath}
\usepackage{amssymb}
\usepackage[ruled]{algorithm2e}

\usepackage[numbers,compress,sort]{natbib}

\title[Hierarchical Knowledge Distillation on Text Graph for Data-limited Attribute Inference]{Hierarchical Knowledge Distillation on Text Graph for Data-limited Attribute Inference}

\author[Q. Li et al.]{%
Quan Li\\
\institute{Pennsylvania State University}\\
\email{qbl5082@psu.edu}\And
Shixiong Jing\\
\institute{Pennsylvania State University}\\
\email{svj5489@psu.edu}\And
Lingwei Chen\\
\institute{Wright State University}\\
\email{lingwei.chen@wright.edu}
}

\begin{document}

\maketitle

\begin{abstract}
The popularization of social media increases user engagements and generates a large amount of user-oriented data. Among them, text data (e.g., tweets, blogs) significantly attracts researchers and speculators to infer user attributes (e.g., age, gender, location) for fulfilling their intents. Generally, this line of work casts attribute inference as a text classification problem, and starts to leverage graph neural networks (GNNs) to utilize higher-level representations of source texts. However, these text graphs are constructed over words, suffering from high memory consumption and ineffectiveness on few labeled texts. To address this challenge, we design a text-graph-based few-shot learning model for attribute inferences on social media text data. Our model first constructs and refines a text graph using manifold learning and message passing, which offers a better trade-off between expressiveness and complexity. Afterwards, to further use cross-domain texts and unlabeled texts to improve few-shot performance, a hierarchical knowledge distillation is devised over text graph to optimize the problem, which derives better text representations, and advances model generalization ability. Experiments on social media datasets demonstrate the state-of-the-art performance of our model on attribute inferences with considerably fewer labeled texts.

\end{abstract}

\section{Introduction}

In the Internet-age, social media has drastically penetrated our everyday lives through countless websites and apps, which allows us to effortlessly connect with each other across the globe, and express personal ideas for social engagements \cite{jia2018attriguard}. Such a convenient environment that brims with vigor and vitality generates a mass of text data reserving basic yet rich user information, which, more importantly, often implies intrinsic user attributes \cite{li2022adversary,li2021turning}, such as age, gender, location, and political view. Due to this fact, different parties have been attracted to reveal user attributes from their text data \cite{li2023adversary}, either conscientiously (e.g., for assessing pandemic risks and analyzing social behaviors \cite{ye2020alpha,lin2020social}) or opportunistically (e.g., for promoting advertisements and tracking users \cite{yu2018adversarial,jia2018attriguard}). 

While the intents of user attribute inferences on social media vary, the methods used to infer such information from text data are consistent. 
Among these developed machine learning models \cite{gong2018attribute,jia2017attriinfer}, natural language processing (NLP) models (e.g., long short-term memory \cite{graves2012long}, and transformer \cite{dai2019transformer}) provide the successful principles to learn high-level representations of source texts.     
Despite the promising performance, their inputs are inherently self-contained, and struggle to leverage structural interactions with other texts. Graph neural networks (GNNs) have recently emerged as one of the most powerful techniques for graph understanding and mining \cite{kipf2016semi,ying2018graph}. These GNNs perform neighborhood aggregations and boost the state-of-the-arts for a variety of downstream tasks over graphs \cite{li2023pseudo,ashmore2023hover,li2023hierarchical}. Therefore, a surge of effective research works apply GNNs to infer user attributes on social media \cite{chen2019semi,mohamed2020social} or simply perform text classification \cite{yao2019graph,ding2020more,wang2021hierarchical,zhang2020every,huang2019text,linmei2019heterogeneous}. For example, Yao et al.\ proposed a GNN-based method to analyze texts by converting the corpus to a heterogeneous graph with words/documents as nodes and word co-occurrence as edges, which requires high memory consumption yet delivers low expression power for individual texts. Huang et al.\ \cite{huang2019text} reduced the computational cost by using global shared word representations, and Ding et al. \cite{ding2020more} defined hyperedges on sequential and topic-related correlations to capture high-order interactions between words. 
Similar refinements can be also found in this line of work \cite{wang2021hierarchical,zhang2020every,linmei2019heterogeneous}. However, different from siamese and matching networks, these GNN-based models construct text graphs simply using local/global word co-occurrence and text-word relations, which may improve the text representations to some extent, but barely work on the application scenarios when labeled texts are few.

Due to privacy concerns, most social media websites and apps limit the access to some personal information; thus, user attribute labels, especially for those private attributes, may only be available on few texts \cite{li2022distilling}. When we reduce user attribute inference problem to text classification problem, we face the challenge that our built model needs to have the ability to learn from few text samples \cite{li2023knowledge}. To address this challenge, we propose a few-shot learning model to implement attribute inferences on social media text data.
Given a text corpus (e.g., tweets, blogs) and an attribute to infer, our model starts by mapping each text to an initial representation; then, a text graph is constructed upon these representations where each node represents one text, 
and edges are learned from the current text representations (either initial ones concatenated with one-hot encoding of attribute label at the input, or hidden representations) via manifold learning. This differs from those static text graphs built upon massive words and offers a better trade-off between expressive power and computational complexity. The task-driven message passing is then conducted directly between labeled and unlabeled text pairs for label propagation, which copes better with data scarcity issue. To further leverage unlabeled texts to improve few-shot performance, a hierarchical knowledge distillation is devised to optimize our graph-based model for attribute inferences: (1) the first level performs on cross domain between source-domain labeled texts and target-domain unlabeled texts to derive better representations, and (2) the second level works on the target labeled and unlabeled texts to advance generalization ability. In summary, our paper has the following major contributions:
\begin{itemize}
    \item We construct text graph via manifold learning to reveal the intrinsic neighborhood among text representations, and refine graph structure via message passing to improve its expressive power and facilitation for label propagation.
    \item We design hierarchical knowledge distillation to utilize both labeled and unlabeled texts for few-shot attribute inference, which first betters text representations from distillation on cross-domain texts, and then advances generalization from distillation on target texts.        
    \item We conduct extensive experiments on real-world social media text datasets with three different attribute settings, which validate that our model can effectively infer user attributes with considerably few labeled texts, and significantly outperforms text-graph baselines.
\end{itemize}

\section{Problem Statement}

In this paper, we put aside the intents (either conscientious or opportunistic) of user attribute inferences, and focus on the investigation of how we can generalize the attribute inference model into a more challenging setting with sparse information on words and few labels on texts, which is more realistic for social media environment.

\begin{figure}[t]
	\centering
	\includegraphics[width=\linewidth]{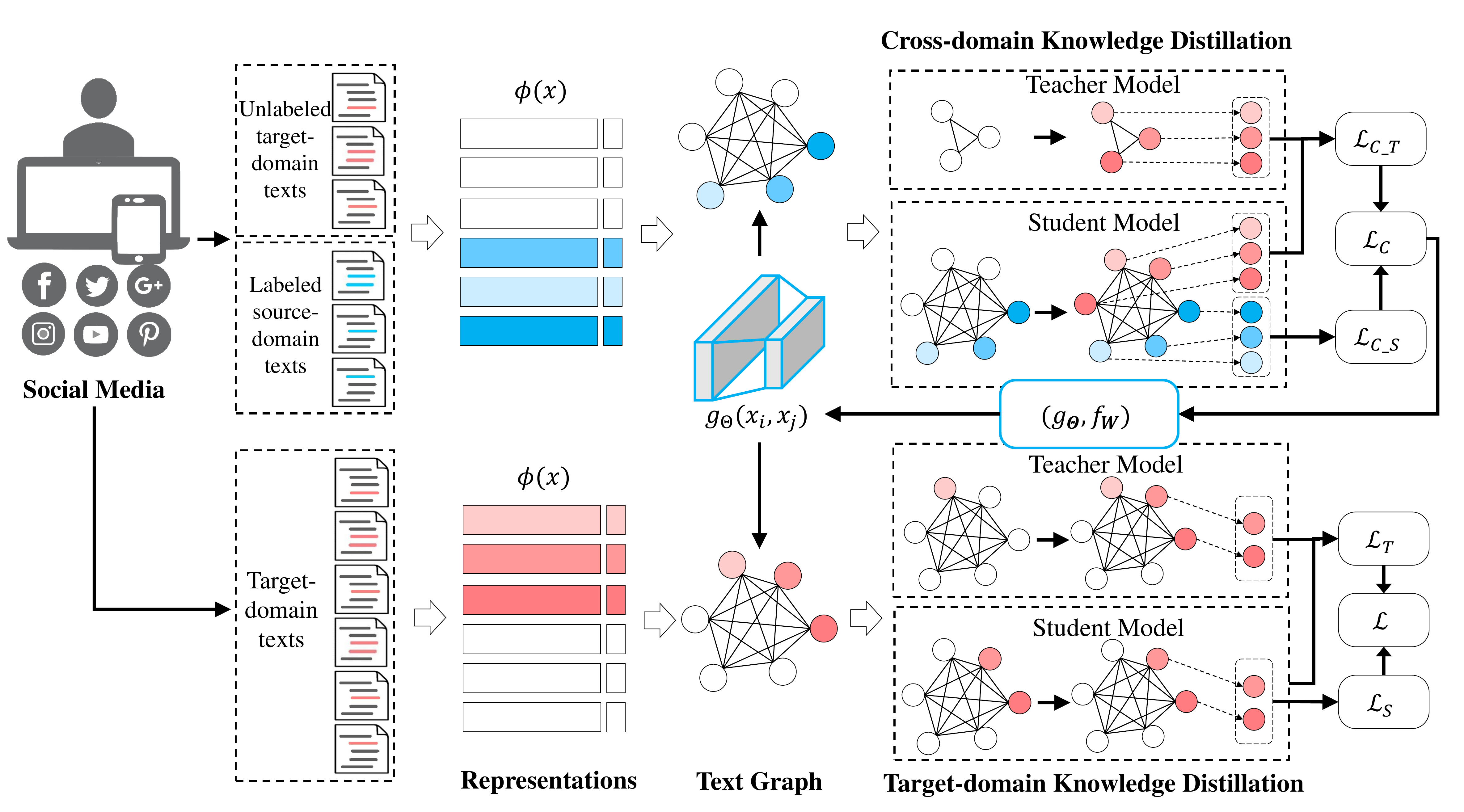}\\
	\caption{The overview of our proposed model, which includes three main components: text representations, text graph construction and refinement, and hierarchical knowledge distillation.} \label{fig:overview}
\end{figure}

Without loss of generality, we represent social media text data as ${\mathcal X} = \{(x_{i}, y_{i})\}_{i = 1}^{m} \cup \{x_{i}\}_{i = 1}^{n}$ consisting of $m+n$ sample texts, where $m$ is the number of the labeled texts and $n$ is the number of unlabeled texts. Unlike existing works \cite{yao2019graph,ding2020more,wang2021hierarchical,zhang2020every,huang2019text,linmei2019heterogeneous} that use sufficient labeled texts for model training, we consider only few of the texts collected from social media have attribute labels, which is the practical scenario. As such, among the social media text data ${\mathcal X}$, $m$ is much smaller than $n$ (i.e., $m \ll n$). Each text $x$ in the labeled text set is annotated with a ground-truth label $y \in {\mathcal Y}$ for a specific attribute. Taking location attribute (main four U.S. regions) as an example: ${\mathcal Y}$ can be accordingly specified as ${\mathcal Y} = \{0\text{:Northeast}, 1\text{:Midwest}, 2\text{:South}, 3\text{:West}\}$. We follow the
general NLP routine to deal with discrete text data by mapping each text $x$ into a $k$-dimensional feature vector $\mathbf x = \phi(x)$ where $\phi$ is a feature representation function $\phi: {\mathcal X} \rightarrow \mathbf X \subseteq \mathbb{R}^{(m+n) \times k}$. Resting on text representations, we aim to learn a text classification model $f: \mathbf X \rightarrow \mathbf Y$ which can take advantage of few labeled texts and large unlabeled texts to perform our social media attribute inference task. Thus, the attribute label of a given text $\mathbf x$ can be inferred using the following formula:
\begin{equation}\label{eq:attributeinference}
    y^{*} = \underset{y \in {\mathcal Y}}{\arg\!\max}\ f_y(\mathbf{x})
\end{equation}
where $f_{y}(\mathbf x)$ is the confidence score of predicting text $\mathbf x$ as attribute label $y$ using the text classification model $f$. From Eq. (\ref{eq:attributeinference}), we can see that the final attribute label assigned to the input text is the one with the highest confidence score. 

\section{Proposed Model}
\label{OurMeth}

In this section, we present the detailed technical steps of how we learn text representations, construct and refine text graph, and how we formulate our text-graph-based few-shot learning model using hierarchical knowledge distillation to perform attribute inferences on social media. The overview of our attribute inference model is illustrated in Figure~\ref{fig:overview}.

\subsection{Text Representations} \label{TextRep}
We aim to build a graph over social media texts directly to benefit few-shot attribute inference. To proceed with graph construction in text granularity, the first step is to initialize each text $x$ into $k$-dimensional feature vector $\mathbf x$ with good expressive quality. Compared to GloVe \cite{pennington-etal-2014-glove}, BERT \cite{devlin2019bert} provides a more context-aware word embedding space, and thus boosts the state-of-the-art performance on the downstream NLP tasks. To this end, we use it to formulate our text representations. More specifically, we leverage SBERT \cite{reimers-2019-sentence-bert} with fine-tuned semantic relations that adds a pooling operation to the output of BERT to derive a fixed-size embedding $\phi_{1}(x)$ for the input text.

In addition, to facilitate label information propagation among labeled and unlabeled nodes via task-driven message passing, we further map the label of each text into a one-hot encoding $\phi_{2}(x)$, and concatenate it with SBERT embedding $\phi_{1}(x)$ as the final text representation at the input of text graph construction, which can be specified as:
\begin{equation}\label{eq:textrepresentation}
    \mathbf x = \phi(x)  = [\phi_{1}(x);\phi_{2}(x)], \mathbf x \in \mathbb{R}^{k}
\end{equation}
Let $\phi_{1}(x) \in \mathbb{R}^{k_1}$ and $\phi_{2}(x) \in \mathbb{R}^{k_2}$ $(k_2 = |{\mathcal Y}|)$; then the dimension of our text representation is $k=k_1 + k_2$. For those texts without attribute labels, we replace the one-hot encoding with the uniform distribution over the $k_2$-simplex, and accordingly get 
\begin{equation}\label{eq:textrepresentation1}
    \phi_{2}(x) = \frac{\boldsymbol{1}_{k_2}}{k_2}
\end{equation}
This formulation to combine SBERT embedding and label encoding as text representation is helpful for our text-graph-based learning model to infer the potential attribute similarity between texts in a data-limited setting.  

\subsection{Text Graph Construction and Refinement}
\label{GraphCons}

The goal of our attribute inference model is to learn from few labeled texts and propagate attribute label information from the labeled texts to the unlabeled ones through their relatedness. Recent researches have demonstrated that message passing with graph-based neural networks can effectively work on such label propagation \cite{garcia2017few,gilmer2017neural,liu2018learning}. In this paper, we extend this paradigm to cast attribute inference using task-driven message passing and infer a text's attribute label from the input texts and labels over text graph. Here, we argue that there are three reasons behind our graph construction over texts rather than word co-occurrences: (1) label propagation can be easily performed as a posterior inference between labeled and unlabeled text pairs, enabling our model to better address labeled data scarcity issue; (2) update on text representations can be immediately used to refine graph structure and improve its expressive power; and (3) as text set is much smaller than word set, the number of graph nodes can be significantly reduced to save the computational cost.

\vspace{0.2cm}\noindent\textbf{Graph construction via manifold learning.}
Given a social media text corpus ${\mathcal X}$, we construct a fully-connected graph $G_\mathcal{X} = (V,E)$ to associate $\mathcal{X}$, where $V$ denotes the set of texts (both labeled and unlabeled), and $E = V \times V$ denotes the set of edges that connect text pairs. Generally, the similarity kernel over node pair is used to build the connection between nodes. Differently, manifold learning\cite{liu2018learning} reveals the low-dimensional manifold embedded in high-dimensional space with the non-linear dimensionality reduction process;
in other words, this can be feasibly exploited to build up the intrinsic neighborhood among text representations. Thus, we initialize each edge $e_{ij}$ between text $v_i$ and text $v_j$ in $G_\mathcal{X}$ by a layerwise non-linear combination of absolute difference between their representations $\mathbf{x}_i$ and $\mathbf{x}_j$ as
\begin{equation}\label{eq:edge}
    e_{ij} = g_{\boldsymbol{\Theta}}(\mathbf{x}_i, \mathbf{x}_j) = \sigma(\cdots\sigma(|\mathbf{x}_i-\mathbf{x}_j|\boldsymbol{\Theta}^{(0)})\cdots\boldsymbol{\Theta}^{(l-1)})\boldsymbol{\Theta}^{(l)}
\end{equation}
where $\sigma(\cdot)$ is a non-linear activation function (e.g., ReLU), and $\boldsymbol{\Theta}$ is learnable weight matrix for each layer. As the constructed structure behaves differently regarding different text representations, the learned edges do not specify a fixed text graph, suggesting the graph can be refined in a discriminative fashion when the neighborhood information is updated.

\vspace{0.2cm}\noindent\textbf{Graph refinement via message passing.}
To refine text graph, we apply iterative message passing through neighborhood structure using a graph convolutional network (GCN) \cite{kipf2016semi,chen2021enhancing} to propagate text features and labels along the labeled and unlabeled nodes, and enhance text representations. Specifically, we build the adjacency matrix $A^{(h)}$ at layer $h$ by normalizing edge matrix using a softmax at each row, where each $e_{ij}$ is computed on the current text representations $\mathbf{x}_i^{(h)}$ and $\mathbf{x}_j^{(h)}$:
\begin{equation} \label{eq:adjacency}
    A_{i,j}^{(h)} = \text{softmax}(g_{\boldsymbol{\Theta}}(\mathbf{x}_i^{(h)}, \mathbf{x}_j^{(h)}))
\end{equation}
Each message passing iteration can be formalized as multi-layer neighborhood information aggregation, which receives current text representation matrix $\mathbf{X}^{(h)}$ as input and produces new text representation matrix $\mathbf{X}^{(h+1)}$ as follows:
\begin{equation} \label{eq:gcn}
    \mathbf{X}^{(h+1)} = \sigma(\widetilde{A}^{(h)}\mathbf{X}^{(h)}\mathbf{W}^{(h)})
\end{equation}
where at layer $h$, $\mathbf{W}$ is weight matrix, $\widetilde{\mathbf{A}} = \mathbf{D}^{-\frac{1}{2}}\hat{\mathbf{A}}\mathbf{D}^{-\frac{1}{2}}$, $\hat{\mathbf{A}} = \mathbf{A} + \mathbf{I}$, and $\mathbf{D}$ is the diagonal degree matrix defined on $\hat{\mathbf{A}}$, i.e., $\mathbf{D}_{ii} = \sum_{j=1}^n \hat{\mathbf{A}}_{ij}$. The text graph $G_\mathcal{X} = (V,E)$ is reconstructed after every message passing iteration by computing each edge as $g_{\boldsymbol{\Theta}}(\mathbf{x}_i, \mathbf{x}_j)$ based on the refined text representations. This gives our text-graph-based attribute inference model more expressive power.

\subsection{Hierarchical Knowledge Distillation}

Our constructed and refined text graph can be used directly to perform posterior inference and propagate the attribute labels from few labeled texts to the target texts via semi-supervised learning, and deliver promising attribute inference performance. In our model formulation, we take a further step to leverage unlabeled texts to improve few-shot learning performance. Specifically, we devise a hierarchical knowledge distillation operation over the text graph to better text representations from knowledge distillation on cross-domain texts, and advance model generalization from knowledge distillation on target texts. The knowledge distillation technique was first designed for model compression, which was then generalized to transfer soft knowledge along teacher neural network to student neural network in a simple way \cite{hinton2015distilling}. Typically, the soft knowledge produced by a neural network is defined as class probabilities output from the softmax layer, where an adjustable temperature parameter controls the final knowledge; a higher temperature produces softer probability distribution over classes. Our hierarchical knowledge distillation operation is detailed as follows.    

\vspace{0.2cm}\noindent\textbf{Cross-domain knowledge distillation.}
The first-level knowledge distillation is performed on cross domain texts. Islam et al. argued that combining cross-domain supervised and unsupervised loss provides better representations for the downstream few-shot learning task \cite{islam2021dynamic}. This inspires us to use knowledge distillation to combine supervised loss from source-domain labeled texts to learn generic text features, and unsupervised loss from target-domain unlabeled texts to develop target-specific text representations. Specifically, let source-domain labeled texts be ${\mathcal X}_{C\_S}$, and target-domain unlabeled texts as ${\mathcal X}_{C\_T_n}$. A teacher model is first trained on target-domain few labeled texts ${\mathcal X}_{C\_T_m}$ to produce pseudo labels for the unlabeled texts as the distilled knowledge:
\begin{equation} \label{eq:cross}
    p(\mathbf{x}_{C\_T_n}|{\mathcal X}_{C\_T_m}) = \frac{\exp{\left(f_{y}(\mathbf{x}_{C\_T_n}/{\tau})\right)}}{\sum_{y\in{\mathcal Y}}\exp{\left(f_{y}(\mathbf{x}_{C\_T_n}/{\tau})\right)}}
\end{equation}
where $\tau$ is distillation temperature, $\mathbf{x}_{C\_T_n} \in {\mathcal X}_{C\_T_n}$, and $f_{y}(\mathbf{x}_{C\_T_n}/{\tau})$ is the confidence score of predicting text $\mathbf{x}_{C\_T_n}$ as attribute label $y$ after iterative message passing over text graph. A student model is then trained on source-domain labeled texts ${\mathcal X}_{C\_S}$ and target-domain unlabeled texts ${\mathcal X}_{C\_T}$. It calculates a cross-entropy loss (supervised loss) between the student's predictions and ground-truth labels, which can be denoted as:
\begin{equation} \label{eq:supervisedloss}
    {\mathcal L}_{C\_S} = -\frac{1}{|{\mathcal X}_{C\_S}|}\sum_{\mathbf{x}_{C\_S}\in {\mathcal X}_{C\_S}} y\log p(\mathbf{x}_{C\_S}|{\mathcal X}_{C}) 
\end{equation}
and a distillation loss (unsupervised loss) between the student's predictions and pseudo labels, which is specified as:
\begin{equation} \label{eq:unsupervisedloss}
    {\mathcal L}_{C\_T} = -\frac{1}{|{\mathcal X}_{C\_T_n}|}\sum_{\mathbf{x}_{C\_T_n}\in {\mathcal X}_{C\_T_n}} p(\mathbf{x}_{C\_T_n}|{\mathcal X}_{C\_T_m})\log p(\mathbf{x}_{C\_T_n}|{\mathcal X}_{C}) 
\end{equation}
where ${\mathcal X}_{C} = {\mathcal X}_{C\_S} \cup {\mathcal X}_{C\_T_n}$. Both the supervised
loss and unsupervised loss are used to learn the student model's weights by optimizing the total loss:
\begin{equation} \label{eq:baseloss}
    {\mathcal L_C} = {\mathcal L}_{C\_S} + {\mathcal L}_{C\_T}
\end{equation}
The trained student model is then used as the base model for the target-domain knowledge distillation in the next step. To take advantage of better text representations derived from cross-domain operation, we only update the weights of the linear layer for classification on the top of the base model, while leaving other parameters unchanged during the final model training.       
  


\begin{algorithm}[t]
	\SetAlgoLined
	\KwIn{${\mathcal X}$: target social media texts with $m$ labeled texts ${\mathcal X}_{C\_T_m}$ and $n$ unlabeled texts ${\mathcal X}_{C\_T_n}$ $(m \ll n)$; ${\mathcal X}_{C\_S}$: source-domain labeled texts; $\phi(\cdot)$: text representation function; $\tau$: distillation temperature; $\lambda$: distillation balance parameter; $T$: epochs.}
	\KwOut{$f$: few-shot attribute inference model.}
	\BlankLine
	$\mathbf{X} = \phi({\mathcal X})$, $\mathbf{X}_{C\_S} = \phi({\mathcal X}_{C\_S})$\;
	// \textbf{Base model training}:\\
	\For{each epoch $t \le T$}
	{
	    Construct and refine $G$ on $\mathbf{X}_{C\_S}$ and $\mathbf{X}_{C\_T_n}$ using Eq.~(\ref{eq:edge}) and Eq.~(\ref{eq:gcn})\;
	    Calculate ${\mathcal L}_{C\_S}$ in Eq.~(\ref{eq:supervisedloss})\;
	    Calculate ${\mathcal L}_{C\_T}$ in Eq.~(\ref{eq:unsupervisedloss})\;
	    Update $\Theta$ and $\mathbf{W}$ by minimizing ${\mathcal L}_{C}$ in Eq.~(\ref{eq:baseloss})\;
	}
	// \textbf{Final model training}:\\
	Load base model $f$ with $\Theta$ and $\mathbf{W}$\;
	Construct and refine $G$ on $\mathbf{X}$\;
	\For{each epoch $t \le T$}
	{
	    Calculate ${\mathcal L}_{S}$ in Eq.~(\ref{eq:studentloss})\;
	    Calculate ${\mathcal L}_{T}$ in Eq.~(\ref{eq:kdloss})\;
	    Update top linear layer of $f$ by minimizing ${\mathcal L}$ in Eq.~(\ref{eq:finalloss})\;
	}
	\caption{Text-graph-based few-shot attribute inference.} \label{alg:fewshotlearning}
\end{algorithm}

\vspace{0.2cm}\noindent\textbf{Target-domain knowledge distillation.}
Based on the trained base model, the second-level knowledge distillation is performed on the target texts to advance the generalization ability of our few-shot learning model for attribute inference. As such, we divide the labeled texts into two categories: teacher texts ${\mathcal X}_{T}$ and student texts ${\mathcal X}_{S}$. A teacher model is trained on ${\mathcal X}_{T}$, which is then used to perform attribute inference on ${\mathcal X}_{S}$. The knowledge distilled by the teacher model can be defined as the inference probability of attribute label for text $\mathbf{x}_{S}$ in ${\mathcal X}_{S}$:
\begin{equation} \label{eq:probability}
    p(\mathbf{x}_{S}|{\mathcal X}_{T}) = \frac{\exp{\left(f_{y}(\mathbf{x}_{S}/{\tau})\right)}}{\sum_{y\in{\mathcal Y}}\exp{\left(f_{y}(\mathbf{x}_{S}/{\tau})\right)}}
\end{equation}
where $\tau$ is the temperature for current-level knowledge distillation, $\mathbf{x}_{S}$ is the representation of the text from ${\mathcal X}_{S}$, and $f_{y}(\mathbf{x}_{S}/{\tau})$ is the confidence score to predict $\mathbf{x}_{S}$ as attribute label $y$ using the base model. Similarly, a student model is trained on ${\mathcal X}_{S}$, which generates inference probability of attribute label for text $\mathbf{x}_{S}$ as $p(\mathbf{x}_{S}|{\mathcal X}_{S})$. Accordingly, the student model may learn the distilled knowledge from the teacher model by optimizing the cross-entropy loss function:
\begin{equation} \label{eq:kdloss}
    {\mathcal L}_T = -\frac{1}{|{\mathcal X}_{S}|}\sum_{\mathbf{x}_{S}\in {\mathcal X}_{S}} p(\mathbf{x}_{S}|{\mathcal X}_{T})\log p(\mathbf{x}_{S}|{\mathcal X}_{S}) 
\end{equation}
$p(\mathbf{x}_{S}|{\mathcal X}_{T})$ is predicted by teacher model on unlabeled data, which can be considered soft attribute label with the same distribution as $p(\mathbf{x}_{S}|{\mathcal X}_{S})$ from student model. This significantly enables the model to learn from unlabeled texts.

\subsection{Loss Generation for Transductive Training}
\label{sec:Loss generation}

The student model itself computes training loss between predictions and ground truth (hard attribute label), which is defined as: 
\begin{equation} \label{eq:studentloss}
    {\mathcal L}_S = -\frac{1}{|{\mathcal X}_{S}|}\sum_{\mathbf{x}_{S}\in {\mathcal X}_{S}} y\log p(\mathbf{x}_{S}|{\mathcal X}_{S}) 
\end{equation}
In this respect, the final objective loss function of our learning model for attribute inference can be formalized as:  
\begin{equation} \label{eq:finalloss}
    {\mathcal L} = (1-\lambda){\mathcal L}_S + \lambda {\mathcal L}_T
\end{equation}
where $\lambda$ is a distillation balance parameter to trade off ${\mathcal L}_S$ and ${\mathcal L}_T$. We train our text-graph-based few-shot learning model in a transductive (or semi-supervised) manner, where all texts (labeled and unlabeled) are accessible during training. Algorithm~\ref{alg:fewshotlearning} illustrates the full steps to leverage hierarchical knowledge distillation on text graph for few-shot attribute inference.

\section{Experimental Results and Analysis}

In this section, we fully evaluate the effectiveness of our proposed text-graph-based few-shot learning model for attribute inference over social media text data and compare it with other baselines. We also investigate the impacts of the hyperparameters and model components on the inference performance.

\subsection{Experimental Setup}

\textbf{Datasets.} We test our model on three real-world social media datasets: GeoText\cite{eisenstein2010latent}, Twitter dataset\footnote{https://www.kaggle.com/crowdflower/twitter-user-gender-classification} and Blog dataset \cite{schler2006effects}, which are  representatives for social media text data. Specifically, GeoText includes the tweets from users with their geographical information. We match users into the regions defined by Census Bureau\footnote{https://www2.census.gov/geo/pdfs/maps-data/maps/reference/us\_regdiv.pdf}, and collect over $9,000$ valid tweets. Due to the data imbalance problem, we choose the two most evenly distributed categories for our experimental evaluation. The Twitter dataset is collected from Kaggle which is composed of tweets, genders and their confidence scores. We filter out those with gender confidence score less than $0.5$, and obtain $13,926$ tweets with two genders (female and male). For Blog dataset, it consists of $19,320$ documents, each of which contains the posts provided by a single user. We extract $25,176$ blogs with two attributes: (1) gender (female and male), and (2) age (teenagers (age between 13-18) and adults (age between 23-45)). Note that, groups with age between 19-22 are missing in the original data. The statistics of these three datasets are summarized in the Table \ref{tab:dataset}.

\vspace{0.2cm}\noindent\textbf{Baselines.}
As our model is built upon text graph, in our comparative study, we select five state-of-the-art text-graph-based models using GNNs to perform text classification tasks and one GNN-based few-shot learning model to be our baselines: 
\begin{itemize}
    \item TL-GNN \cite{huang2019text}: It learns a global shared word representations for the whole dataset and builds a graph on the basis of word embeddings in the documents, where message passing is used for text classification.
    \item HyperGAT \cite{ding2020more}: It defines two types of hyperedges to link word tokens in documents while constructing graph, based on which it trains a graph model by using a dual-attention mechanism to aggregate neighborhood information.
    \item TextGCN \cite{yao2019graph}: It considers both words and documents as nodes in text graph, which is one of the most efficient methods in the early studies. But it does not consider the relations between different documents. 
    \item TextING \cite{zhang2020every}: It builds an individual graph for each text document and uses a gated GNN model to learn word embeddings for the classification task.
    \item HGAT \cite{linmei2019heterogeneous}: It extracts words, documents, topics, and entities as different types of nodes and constructs a heterogeneous graph for the text data. To aggregate information more accurately, it also assigns different importance to different edges based on node types during message passing.
    \item TPN \cite{liu2019fewTPN}: It is a few-shot learning model which deals with node classification with GNN. We replace the original node embeddings that derive from CNN with text representations and test its performance on few-shot text classification task.
\end{itemize}

\begin{table}[t]
    \centering
    \caption{\label{tab:dataset}Comparing statistics of the two datasets}
    \tabcolsep=6pt
    \begin{tabular}{ccccc}
    \toprule
          \textbf{Dataset} & \textbf{Attribute} & \textbf{\#Post} & \textbf{\#Class} & \textbf{\#Vocabulary}\\
          Twitter & Gender & 13,926 & 2  & 21k\\
          Blog & Gender, Age & 25,176 & 2 & 30k\\
          GeoText & Location & 9,290 & 4 & 26k\\
    \bottomrule
    \end{tabular}
\end{table}


\vspace{0.2cm}\noindent\textbf{Parameter setting.} The parameters used to perform hierarchical knowledge distillation and few-shot attribute inferences are specified as follows: 
\begin{itemize}
\item 
Cross-domain training for the base model: To perform cross-domain knowledge distillation, we set blog-age dataset as source-domain texts for blog-gender, twitter-gender, and twitter-location attribute inference tasks, while for blog-age inference task, we use twitter-gender as the labeled source-domain texts. We pre-train attribute inference model using few labeled texts as the teacher model to compute distillation loss for each inference task, and set the distillation temperature $\tau = 3$ to learn the knowledge from the teacher model.

\item Target-domain training for the final model: We randomly select 15 labeled instances per class as training data and select 20\% instances from all the remaining as test data for each inference task. We set the knowledge distillation temperature $\tau = 3$ and the balance parameter $\lambda = 0.3$ for the training loss. We also evaluate the impacts of training size, distillation temperature, and distillation balance parameter in Section~\ref{subsec:evaluation}. 
\end{itemize}

\subsection{Evaluation of Our Model}\label{subsec:evaluation}

\begin{figure}[t]
	\centering
	\begin{tabular}{c c c}
	    \hspace{-0.5cm}
		\includegraphics[clip,trim=0 0 20 30,width=0.34\linewidth]{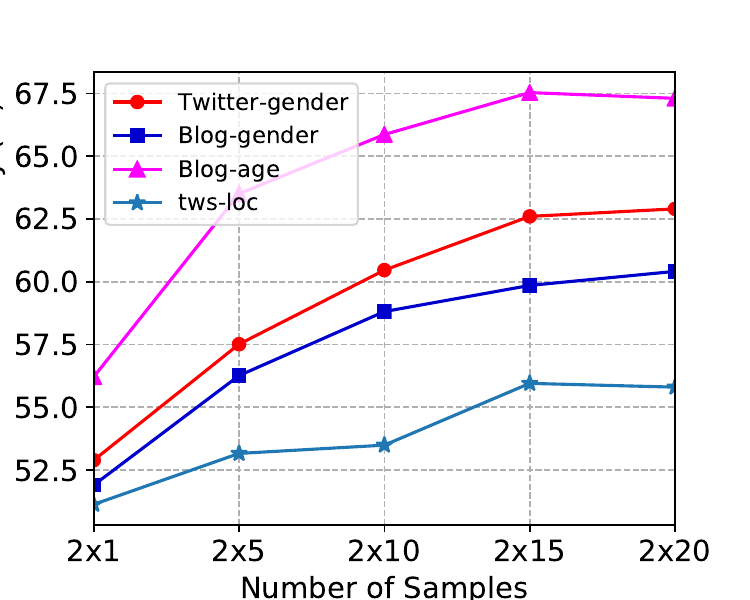}&
		\hspace{-0.5cm}
		\includegraphics[clip,trim=0 0 20 30,width=0.34\linewidth]{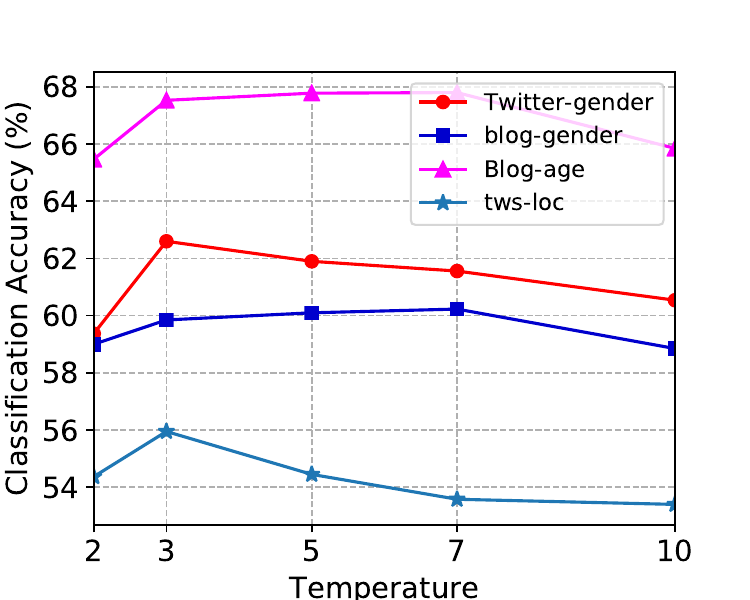}&
		\hspace{-0.6cm}
		\includegraphics[clip,trim=0 0 20 30,width=0.34\linewidth]{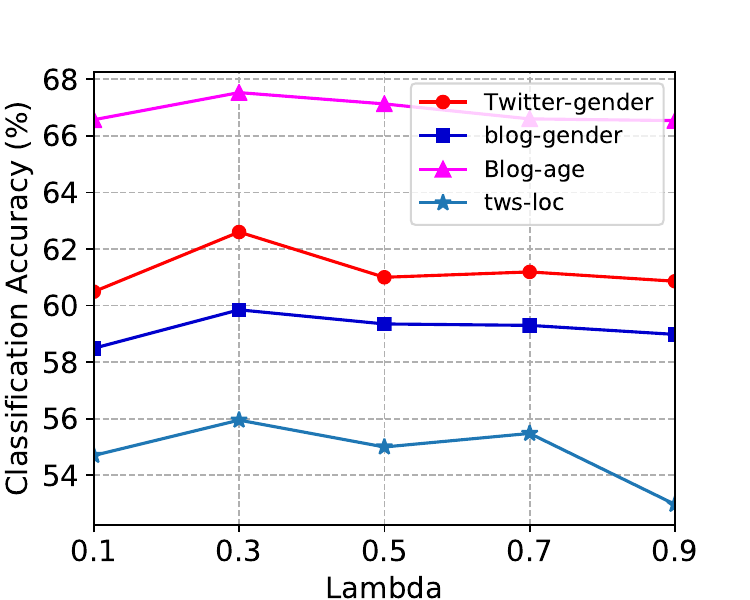}\\
		\hspace{0.2cm}\textbf{(a)} & \hspace{0.2cm}\textbf{(b)}& \hspace{0.2cm}\textbf{(c)}\\
	\end{tabular}
	\caption{Evaluation on different model parameters: (a) sizes of training samples $m$, (b) distillation temperatures $\tau$, and (c) distillation balance parameter $\lambda$.} \label{fig:evaluate}
	\vspace{-0.5cm}
\end{figure}

\noindent\textbf{Effectiveness.}
In this section, we evaluate the effectiveness of our model over three inference settings under different parameters. In particular, we test the inference accuracy of our model with training size $m \in \{2\times1, 2\times5, 2\times10, 2\times15, 2\times20\}$ respectively, while the knowledge distillation temperature $\tau \in \{2, 3, 5, 7, 10\}$ and distillation balance parameter $\lambda \in \{0.1, 0.3, 0.5, 0.7, 0.9\}$ when $m = 2\times15$. 
The experimental results are shown in Figure~\ref{fig:evaluate}. As we can see, though different parameters contribute to different test results, which will be discussed later, our model achieves the state-of-the-art results of inferring attributes on social media texts when only few labeled texts are available. When ``1-shot'' ($2\times1$) is set, the inference accuracy is 52.89\%, 51.90\%, 56.21\%, and 51.13\% for Twitter-gender, Blog-gender, Blog-age, and Twitter-location separately, which are either outperforming or comparable to the performance of the most baselines trained on ($2\times15$); averagely, their inference accuracies are 58.96\%, 57.45\%, 64.08\%, and 53.91\%.    

\begin{table*}[t]
\centering
    \caption{Comparisons of different graph-based baselines ($2 \times 15$)}
    \label{tab:baselines}
    \tabcolsep=5.5pt
    \begin{tabular}{lcccccccc}
    \toprule
       \multirow{2}{*}{\textbf{Inference}} &\multicolumn{2}{c}{Twitter-Gender} &\multicolumn{2}{c}{Blog-gender}  &\multicolumn{2}{c}{Blog-age} &\multicolumn{2}{c}{Twitter-location}\\
       \cmidrule{2-9}
       & ACC(\%) & F1 & ACC(\%) & F1 &ACC(\%) & F1 & ACC(\%) & F1\\
        \midrule
         TL-GNN & 50.49 & 0.3616 & 51.26 & 0.3636 & 56.10 & 0.4282 & 49.68 & 0.3206 \\
         HyperGAT & 50.76 & 0.4755 & 51.88  & 0.3487 & 47.43  & 0.4666 & 50.81 &  0.4968\\
         TextGCN & 49.36 & 0.4314 & 53.43  & 0.5302 & 52.30  & 0.5209 & 52.98 & 0.4882\\
         TextING & 51.36 & 0.4898 & 52.76  &  0.5110 & 58.28  & 0.5696 & 50.45 & 0.4893\\
         HGAT & 52.41 & 0.3439 & 51.69  &  0.3407 & 58.56 & 0.4770 & 47.67 & 0.4259 \\
         TPN & 55.20 & 0.5355 & 52.17 & 0.3876 & 53.20 & 0.3828 & 51.20 & 0.3984 \\
         Our Model & \textbf{62.60} & \textbf{0.5884} & \textbf{59.85} & \textbf{0.5664} & \textbf{67.53} & \textbf{0.6371}  & \textbf{55.95} & \textbf{0.5201} \\
    \bottomrule
    \end{tabular}
\end{table*}




\vspace{0.2cm}\noindent\textbf{Impact of training size $m$.}
As illustrated in Figure~\ref{fig:evaluate}(a), when ``higher-shot'' is applied in training, the performance of our model generally continues to improve, but the improvements of few-shot learning in $[2\times 10, 2\times 20]$ are less significant (or more stable) than that of $[2\times 1, 2\times 10]$. With the training size increases, the advantage of our few-shot model narrows since more labeled texts are used and the inference performance is closer to the upper bound. 

\vspace{0.2cm}\noindent\textbf{Impact of distillation temperature $\tau$.}
As for the distillation temperature, Figure~\ref{fig:evaluate}(b) indicates that when we enlarge $\tau$, the attribute inference accuracy first significantly increases, peaks at $\tau = 3$, then either stays flat or drastically decreases when $\tau$ keeps increasing.
The trend is understandable: when $\tau$ is relatively small, the soft attribute label probabilities distilled from teacher model are informative and helpful to facilitate optimizing student model; when $\tau$ is large, the distilled information from teacher model is more ambiguous, which may in turn smooth the student model's inference ability.

\vspace{0.2cm}\noindent\textbf{Impact of distillation balance parameter $\lambda$.}
As shown in Figure~\ref{fig:evaluate}(c), the inference accuracy rises up when increasing the value of distillation balance parameter $\lambda$ and peaks at $\lambda=0.3$; then it trends to drop slightly after $\lambda=0.5$. The reason for this tendency is that when the value of $\lambda$ increases, the student model learns more knowledge from the teacher model with respect to the soft label and less from the ground truth label, which may first benefit the student model's generalization ability and then likely make it smooth with large $\lambda$. Another observation from Figure~\ref{fig:evaluate}(c) is that blog-age inference setting seems less sensitive to $\lambda$ than others. 

%


\subsection{Comparisons with Baselines}
In this section, we compare our model with five GNN-based baselines that work on text classification over graph structure and one GNN-based few-shot learning baseline, including TL-GNN \cite{huang2019text}, HyperGAT \cite{ding2020more}, TextGCN \cite{yao2019graph}, TextING \cite{zhang2020every}, HGAT \cite{linmei2019heterogeneous}, and TPN\cite{liu2019fewTPN}. The comparative results are illustrated in Table~\ref{tab:baselines} with $m=2\times15$. We can observe that among baselines, HGAT, TextGCN, TextING, and TPN slightly take the lead in Twitter-gender, Blog-gender, Blog-age, and Twitter-location respectively with respect to accuracy and F1-score. It is obvious that our model completely outperforms baselines with a large margin in lower-shot (i.e., the improvement margin of accuracy is $(6.42, 20.10)\%$, and the improvement margin of F1-score is $(0.05, 0.22)$). Another observation from Table~\ref{tab:baselines} and Figure~\ref{fig:evaluate}(a) is that our model with only 1-shot is either outperforming or comparable to baselines with 15-shot. This confirms that (1) graphs built upon word co-occurrence can improve text representations, but hardly learn from few labeled texts; (2) the text-level graph with neighborhood refinement contributes better to few-shot learning than the word-level graph, and (3) our model offers a better trade-off between expressive power and complexity in terms of node number, and thus provides a better solution for social media attribute inferences.

\subsection{Ablation Study}
In this section, we conduct the ablation study to further investigate how different components contribute to the performance of our model. Our model proceeds with text representations, graph construction and refinement, and two-level knowledge distillations. We gradually add these components one by one and formulate five attribute inference models: (1) SBERT: directly feed SBERT representations to fully-connected and softmax layers for text classification; (2) SBERT+Graph: construct and refine a text graph using SBERT representations and perform posterior inference through transductive learning; (3) SBERT+Graph+KD\_1: apply the first-level knowledge distillation to leverage cross-domain information; (4) SBERT+Graph+KD\_2: apply the second-level knowledge distillation to leverage target-domain information; (5) SBERT+Graph+KD\_1+KD\_2: the complete design of our model. The results are reported in Table \ref{tab:ablationstudy}.


\begin{table}[t]
    \tiny
    \centering
    \caption{Evaluation on model components (accuracy \%)} \label{tab:ablationstudy}
    \resizebox{\linewidth}{!}{%
    \begin{tabular}{|c|c|c|c|c|c|c|c|}
        \hline
         SBERT & Graph & KD\_1 & KD\_2 & Twitter-gender  & Twitter-location & Blog-gender & Blog-age \\
         \hline
         \checkmark & & & & 51.20 & 50.17 & 50.93 & 54.59  \\
         \checkmark & \checkmark & & & 58.04 & 53.43 & 55.35 & 64.64 \\
         \checkmark & \checkmark & \checkmark &  & 60.68 & 54.20 & 58.44 & 66.53 \\
         \checkmark & \checkmark & & \checkmark & 61.16 & 54.35 & 59.05 & 66.80 \\
         \checkmark & \checkmark & \checkmark & \checkmark & \textbf{62.60} & \textbf{55.95} & \textbf{59.85} & \textbf{67.53} \\
         \hline
    \end{tabular}
   
    } \vspace{-0.4cm}
\end{table}


As we can see from Table \ref{tab:ablationstudy}, SBERT representations provide good expressive quality for texts, which achieve comparable performances to some baselines over word-level graphs, since those text representations learned from word-level graphs barely consider the contextual correlations within texts. The constructed and refined text graph learned through manifold learning and message passing plays an important role to the efficacy of our model. With this component added, the inference accuracy significantly increases by $(3.0, 11.0)\%$. 
When first-level and second-level knowledge distillation is performed individually, the inference model derives better text representations by combining supervised and unsupervised loss from cross-domain and target-domain texts respectively, which improves inference accuracy by $(1.0, 4.0)\%$. The hierarchical knowledge distillation aggregating two-level information is able to further advance the state-of-the-art performance to a higher level, which implies that this operation yields an additional advantage for few-shot learning. These observations reaffirm the effectiveness of our design to infer attributes on social media when labeled texts are few.

\vspace{-0.4cm}
\section{Conclusion}
In this work, we investigate social media attribute inferences in the more challenging and practical setting with sparse information on words and few labels on texts. More specifically, we design a text-graph-based few-shot learning model to address this challenge. In particular, we use manifold learning and message passing to construct and refine the text-graph to offer a better trade-off between expressive power and computational complexity. And then, we devise a hierarchical knowledge distillation operation over the text graph to better text representations from knowledge distillation on cross-domain texts, and advance model generalization ability from knowledge distillation on target texts. To evaluate the effectiveness of our designed model, we conduct extensive experiments on 
three real-world social media datasets and three realistic inference settings. The state-of-the-art results demonstrate the effectiveness of our model in the challenging few-shot setting for attribute inferences, and validate its superiority to baselines. In addition, we reveal that our model provides great value and general validity for attribute inference in practice. 

\bibliographystyle{unsrtnat}
\bibliography{reference}

\end{document}